\documentclass[11pt,a4paper, final, twoside]{article}
\usepackage{amsmath}
\usepackage{fancyhdr}
\usepackage{amsthm}
\usepackage{amsfonts}
\usepackage{amssymb}
\usepackage{amscd}
\usepackage{latexsym}
\usepackage{amsthm}
\usepackage{graphicx}
\usepackage{graphics}
\usepackage{afterpage}
\usepackage[colorlinks=true, urlcolor=blue,  linkcolor=black, citecolor=black]{hyperref}
\usepackage{color}
\usepackage[table,xcdraw]{xcolor}
\usepackage{float}
\usepackage[linesnumbered,ruled,vlined, algo2e]{algorithm2e}

\SetCommentSty{mycommfont}

\SetKwInput{KwInput}{Input}                
\SetKwInput{KwOutput}{Output}              

\theoremstyle{proposition}
\newtheorem{prop}{Proposition}[section]

\theoremstyle{definition}

\theoremstyle{remark}

\numberwithin{equation}{section}
\begin{document}
\hyphenpenalty=100000

\setlength{\arrayrulewidth}{0.4mm}
\setlength{\tabcolsep}{12pt}
\renewcommand{\arraystretch}{1.5}

\begin{flushright}

{\Large \textbf{\\Meta-Heuristic Solutions to a Student Grouping
Optimization Problem faced in Higher
Education Institutions }}\\[5mm]
{\large \textbf{Kenekayoro Patrick$^\mathrm{*1}$ and Biralatei Fawei$^\mathrm{1}$\footnote{\emph{*Corresponding author: E-mail: Patrick.Kenekayoro@outlook.com}}  }}  \\[1mm]
$^\mathrm{1}${\footnotesize \it Department of Mathematics and Computer Science, Niger Delta University, \\ Amassoma, Bayelsa State,
Nigeria.}
\end{flushright}

{\Large \textbf{Abstract}}\\[4mm]
\fbox{%
\begin{minipage}{5.4in}{\footnotesize 

Combinatorial problems which have been proven to be NP-hard are faced in Higher Education Institutions and researches have extensively investigated some of the well-known combinatorial problems such as the timetabling and student project allocation problems.  However, NP-hard problems faced in Higher Education Institutions are not only confined to these categories of combinatorial problems.  The majority of NP-hard problems faced in institutions involve grouping students and/or resources, albeit with each problem having its own unique set of constraints. Thus, it can be argued that techniques to solve NP-hard problems in Higher Education Institutions can be transferred across the different problem categories. As no method is guaranteed to outperform all others in all problems, it is necessary to investigate heuristic techniques for solving lesser-known problems in order to guide stakeholders or software developers to the most appropriate algorithm for each unique class of NP-hard problems faced in Higher Education Institutions. To this end, this study described an optimization problem faced in a real university that involved grouping students for the presentation of semester results. Ordering based heuristics, genetic algorithm and the ant colony optimization algorithm implemented in Python programming language were used to find feasible solutions to this problem, with the ant colony optimization algorithm performing better or equal  in 75\% of the test instances and the genetic algorithm producing better or equal results in 38\% of the test instances.

} \end{minipage}}\\[1mm]

\section{Introduction}\label{I1}
NP-hard problems, among which include the University Course Timetabling Problem, Examination Timetabling
Problem and Student Project Allocation Problem are faced annually in Higher Education Institutions. These problems
and strategies to tackle them efficiently have been researched extensively. However, there is a dearth of research on a
combinatorial problem that involves grouping students for the presentation of semester results.

In some Higher Education Institutions, it is necessary to present the examination results of students in a programme. The
document presented in a tabular form usually contains students in rows and courses as column headers, where the value
of Table(i,j) is the examination score of student i in course j.
Given that students register/write exams for different courses and there is only a fixed number of courses that could
appear as column headers for the result presentation document to be legible, it becomes necessary to group students into
subsets such that the union of the courses students took exams for in each group is not more than the number required
for the creation of a legible document. As each group will have a unique set of course headers and may need to start on
a new page in the document, the argument is that fewer groups will need fewer pages to be presented which will save
resources for institutions that store printed copy of result summaries and also result in smaller sizes of the documents in
electronic form. The Student Result Grouping (SRG) problem in its simplest form is thus to find the student groupings
such that the number of groups is minimized. Institutions may have additional constraints and the Student Result
Grouping problem from the university used as the case study in this research is described in the next section.

The SRG problem in the case study university has been solved by greedily assigning or creating groups based on the order students were stored in the database, but this research aims to use heuristic techniques such as the genetic algorithm and the ant colony optimization algorithm to find more suitable solutions to this problem.

\section{Related Work}\label{LIT-REVIEW}

The Student Result Grouping Problem in this study, albeit with its unique requirements is a subset of previously researched NP-hard problems faced in Higher Education Institutions. The University Course Timetabling Problem (UCTP) and Examination Timetabling Problem (ETP) are arguably the most popular of this category of NP-hard problems. The UCTP and ETP involve scheduling events in a Higher Education Institution based on available resources(rooms or facilities) and student or staff specific requirements. 

Meta-heuristics techniques such as Tabu Search \cite{Abdullah2012, Lu2010}, Simulated Annealing \cite{Abdullah2010}, Ant Colony Optimization \cite{Kenekayoro2016} have been investigated for solving these NP-hard problems. \cite{Babaei2015} exhaustively analysed the approaches among others that have been used successfully used to find optimal solutions to these problems. 

The techniques \cite{Babaei2015} reviewed can be transferred to solving other grouping problems faced in Higher Institutions that such as the Student Problem Allocation Problem have been shown to be NP-hard \cite{Manlove2020}. For example, Manlove et al. \cite{Manlove2020} has solved a Student Project Allocation Problem with Integer Programming and Kenekayoro et al.  \cite{kenekayoro2020population} found optimal solutions with Ant Colony Optimization, Genetic Algorithm and the Gravitational Search Algorithm.

Hence, the Student Grouping Problem can be seen as a classic NP-hard problem that can be solved using meta-heuristic techniques that have been researched extensively. This research reports on using the Genetic Algorithm, Ant Colony Optimization and an ordering based heuristic to solve a unique grouping problem faced a case study Higher Education Institution.  

\section{Student Result Grouping (SRG) Problem}\label{I2-SRGP}

A dataset containing a list of students with courses they took examinations for as shown in Table \ref{tab:1} and a list of courses
with the year the courses were introduced to students as shown in Table \ref{tab:2} represents the information available to
produce a suitable grouping.

\begin{table}[h]
\caption{List of students with the courses the students registered}\label{tab:1}
\small
\centering
\begin{tabular}{p{6cm}p{1.5cm}}
\hline
\textbf{Student} & \textbf{Course} \\
\hline
11100011 & CMP301 \\
11100011 & MTH101 \\
11200012 & CMP436 \\
11200012 & CMP421 \\
\hline
\end{tabular}
\end{table}

\begin{table}[h]
\caption{Courses and the year the courses are introduced to students}\label{tab:2}
\small
\centering
\begin{tabular}{p{6cm}p{1.5cm}}
\hline
\textbf{Student} & \textbf{Course} \\
\hline
CMP301 & $3^{rd}$ \\
MTH101 & $1^{st}$ \\
CMP436 & $4^{th}$ \\
CMP421 & $4^{th}$ \\
\hline
\end{tabular}
\end{table}

Assuming that we are to generate groupings for fourth-year students, new courses are fourth-year courses while other courses are categorised as old. In the case study university, students are not allowed the register or take examinations for courses they have not been introduced to. Consequently, fourth-year students can register for any course but third-year students can register for only first, second and third-year courses.

After examinations, the student grouping problem involves grouping students in a programme such that:

\begin{enumerate}
  \item All students are assigned to a group.
  \item The number of unique courses taken by all students in each group is less than 26.
  \item The number of unique new courses taken by all students in each group is less than 13.
  \item The number of unique old courses taken by all students in each group is less than 13.
\end{enumerate}

The numbers in the objectives above were determined to be the most appropriate for legible presentation of results in an A4 sheet by the case study institution.

fourth-year students in the case study university can register courses from any level, thus presenting results for this set of students is a harder problem compared to students at lower levels.

\subsection{Mathematical Formulation}
The Student Result Grouping Problem is made up of a set of students $\lbrace S_1~\cdots~S_m \rbrace$, set of courses $\lbrace C_1~\cdots~C_n \rbrace$ and set of years $\lbrace Y_1~\cdots~Y_o \rbrace$; $Y$ being the years a student can spend in the university ($Y_1 = first~year~students$ and $Y_o = final~year~students$). 

Students are grouped according to the number of years currently spent in the university $SY_j \in \lbrace S_1~\cdots~S_m \rbrace$ and each course $C_i \in \lbrace C_1~\cdots~C_n \rbrace$ is introduced to students $SY_j$ in a particular year $Y_j \in \lbrace Y_1~\cdots~Y_o \rbrace$. 

Exams are taken for Courses. $CY_j$ is the set of courses introduced to students $SY_j$ in year $Y_j$. Students are not allowed to take exams for courses they have not been introduced to. 

For each student $S_k \in SY_j$, course $C_i$ is classified as old if $C_i \notin CY_j$, otherwise ($C_i \in CY_j$), it is classified as new.  

The Student Result Grouping Problem groups all students in a particular year ($SY$) to subgroups $SG$. $SG$ is a set of groups $\lbrace SG_1~\cdots~SG_k \rbrace$ with each group in $SG$ a subset of the students in $SY$; $SG \in SY$ and $CG$ the courses students in $SG$ took exams for. The Student Result Grouping Problem groups students whilst ensuring that:
\begin{enumerate}
  \item All students are assigned to a group. \\
        $ SG_1 \cup SG_2 \cup \cdots \cup SG_k = SY$
  \item No student is assigned to more than one group \\
        $n(SG_1) + n(SG_2) + \cdots + n(SG_k) = n(SY)$
  \item The number of unique courses taken by all students in each group is less than 26 \\
        $n(CG) < 26$
  \item The number of unique new courses taken by all students in each group is less than 13 \\
        $n(CG; ~if~CG_i~is~classified~as~new) < 13 $
  \item The number of unique old courses taken by all students in each group is less than 13 \\
        $n(CG; ~if~CG_i~is~classified~as~old) < 13 $
\end{enumerate}

\subsection{Data}\label{I3}
The dataset downloadable from \href{https://doi.org/10.6084/m9.figshare.12116667}{https://doi.org/10.6084/m9.figshare.12116667} is in the format shown in Table \ref{tab:3-Example-Dataset} with each row showing a student, a course and the year the course was introduced to the student.

\begin{table}[h]
\caption{An example instance in the dataset available from \href{https://doi.org/10.6084/m9.figshare.12116667}{https://doi.org/10.6084/m9.figshare.12116667} }\label{tab:3-Example-Dataset}
\small
\centering
\begin{tabular}{
>{\columncolor[HTML]{EFEFEF}}l 
>{\columncolor[HTML]{EFEFEF}}l 
>{\columncolor[HTML]{EFEFEF}}l }
\hline
\multicolumn{1}{|l}{\cellcolor[HTML]{EFEFEF}\textbf{Student}} & \textbf{Course} & \multicolumn{1}{l|}{\cellcolor[HTML]{EFEFEF}\textbf{Year}} \\ \hline
\multicolumn{1}{|l}{\cellcolor[HTML]{EFEFEF}111011}    & CMP201           & \multicolumn{1}{c|}{\cellcolor[HTML]{EFEFEF}2}             \\
\multicolumn{1}{|l}{\cellcolor[HTML]{EFEFEF}111011}    & CMP421           & \multicolumn{1}{c|}{\cellcolor[HTML]{EFEFEF}4}             \\
\multicolumn{1}{|c}{\cellcolor[HTML]{EFEFEF}...} & \multicolumn{1}{c}{\cellcolor[HTML]{EFEFEF}...} & \multicolumn{1}{c|}{\cellcolor[HTML]{EFEFEF}...}           \\
\multicolumn{1}{|c}{\cellcolor[HTML]{EFEFEF}...} & \multicolumn{1}{c}{\cellcolor[HTML]{EFEFEF}...} & \multicolumn{1}{c|}{\cellcolor[HTML]{EFEFEF}...}           \\
\multicolumn{1}{|c}{\cellcolor[HTML]{EFEFEF}...} & \multicolumn{1}{c}{\cellcolor[HTML]{EFEFEF}...} & \multicolumn{1}{c|}{\cellcolor[HTML]{EFEFEF}...}           \\

\multicolumn{1}{|l}{\cellcolor[HTML]{EFEFEF}303101}             & CMP401           & \multicolumn{1}{c|}{\cellcolor[HTML]{EFEFEF}4}             \\ \hline
\end{tabular}
\end{table}

Table \ref{tab:4-Des.Instances} shows the description of instances in the dataset, which shows an overview of fourth-year students’ examination/course registrations from the Computer Science department of the case study institution. 

\begin{table}[h]
\caption{Description of instances in the Student Result Grouping dataset}\label{tab:4-Des.Instances}
\tiny
\centering
\begin{tabular}{lccc}
\hline
Name     & New Courses & Old Courses & No. of Students  \\
\hline
RGD41107 & 28          & 15          & 140              \\
RGD4152  & 13          & 20          & 97               \\
RGD4185  & 27          & 20          & 121              \\
RGD42118 & 8           & 0           & 25               \\
RGD4263  & 19          & 20          & 132              \\
RGD4296  & 23          & 17          & 193              \\
RGD41118 & 10          & 1           & 18               \\
RGD4196  & 26          & 17          & 185              \\
RGD4241  & 8           & 21          & 67               \\
RGD4274  & 20          & 23          & 147              \\
RGD4141  & 9           & 17          & 68               \\
RGD4174  & 29          & 19          & 149              \\
RGD42107 & 26          & 16          & 123              \\
RGD4252  & 9           & 24          & 95               \\
RGD4285  & 19          & 21          & 128              \\
RGD4163  & 26          & 17          & 128              \\
\hline
\end{tabular}
\end{table}

\section{Methods}\label{I4}
Heuristic algorithms have successfully been used to solve NP-hard problems such as the Student Project Allocation Problem \cite{kenekayoro2020population, chiarandini2018handling, Kwanashie2015} and the Nurse Rostering Problem \cite{Awadallah2013, Jaradat2019}.

Heuristic techniques that find satisfactory solutions to the Examination and Course timetabling problems have also been extensively investigated \cite{Li2010, Tarawneh2011, Kenekayoro2016, Gulcu2020}. Implementation of some of these algorithms are available as packages, for example, DEAP - Distributed Evolutionary Algorithms in Python \cite{Fortin2012} is a python implementation of the Genetic algorithm. This study explores the use of some well-known algorithms to solve the combinatorial problem in this research and compares the quality of resulting solutions. 

\subsection{Evaluation/Fitness function}\label{I5}
The fitness or evaluation function is arguably the most important aspect of a heuristic algorithm. It should be a function that not only evaluates the quality of a solution but also guides an algorithm towards optimal solutions. For example, in the fitness function in this study, the penalty for groups that exceed the required number of unique courses is higher in a group with more students compared to a group with fewer students. The reason being that reducing the number of students in a group can guide the algorithm to a solution with a fewer required number of unique course violations (unfit penalty). 

Three penalties, unfit penalty, unassigned penalty and size penalty are used in the fitness function to determine the quality of a solution to the student result grouping problem in this research.

\begin{prop}The \textbf{unfit penalty} evaluates if the number of unique new and old courses (cardinality of the set of new and old) offered by students grouped together exceeds the allowed limit (13). This is computed with equations \ref{equ.1} and \ref{equ.2}
\begin{equation} 
gp(old|new) = \left\{\begin{matrix}
n(CG_{old|new}) - 13 & if~n(CG_{old|new}) \geq 13 \\ 
0 & otherwise 
\end{matrix}\right.
\label{equ.1}
\end{equation}

\begin{equation} 
up = \sum_{i=1}^{m}\left ( gp_i(new) + gp_i(old) \right ) \times n(SG_i); m~is~the~no.~of~student~groups
\label{equ.2}
\end{equation}

\label{prop.1}\end{prop}

Equation \ref{equ.1} is used to compute the penalties of new and old courses in groups, while the unfit penalty is determined by the sum of group penalties for all groups in the solution as shown in Equation \ref{equ.2}.

\begin{prop} The \textbf{size penalty} as shown in Equation \ref{equ.3} guides algorithms to better solutions by ensuring that the meta-heuristic algorithm favours larger groups instead of smaller groups.
\begin{equation} 
sp = \sum_{i = 1}^{m} (n(SY) - n(SG_i)) \times n(SG_i); m~is~the~no.~of~student~groups
\label{equ.3}
\end{equation}
  
\label{prop.2}\end{prop}

\begin{prop} The \textbf{unassigned penalty} penalizes solutions that do not assign groups to all students. It is simply
evaluated as the number of students not assigned to a group.  
\begin{equation} 
ap = n(SY) - \sum_{i = 1}^{m} (n(SG_i)); m~is~the~no.~of~student~groups
\label{equ.3a}
\end{equation}
\label{prop.4}\end{prop}

The final fitness function is the weighted sum of individual penalties as shown in Equation \ref{equ.4}.

\begin{equation} 
f(x) = 1000 \times ap + (log_2 (sp) + up \times 1000 ) \times n(SG)
\label{equ.4}
\end{equation}

The unfit penalty favours smaller groups because fewer students in a group result in fewer unique courses while the
size penalty favours larger groups. As the size penalty is not as a result of a constraint violation, but help guide heuristic
algorithms to a solution with a fewer number of groups, its weight is scaled-down compared to the unfit penalty, ensuring that meta-heuristic algorithms will try to reduce the number of groups only when the unique courses in all groups do not exceed
the allowed limit.

\subsection{Hardest First Ordering Heuristics}
This is a graph-based technique that has been employed in solving a number of NP-hard problems faced in Higher
Education Institutions (HEIs), particularly the Education Timetabling Problem. Early techniques \cite{Qu2009, Burke2010} that solved timetabling problems ordered events to be scheduled based on the number of constraint
violations scheduling that event may cause, or how difficult scheduling that event is, in terms of student or lecturer
clashes. For example, using the largest degree first graph heuristic, events with more common students are scheduled
first while the least saturated degree schedules events with fewer available timeslots first \cite{Burke2010}.

In the university timetabling problem, the largest degree first and least saturated degree heuristics are variations of the
hardest first heuristic, with difference only in how the hardest event to be scheduled, is determined; largest degree first
uses events with a higher number of student groups, while least saturated degree first uses events with fewer available
timeslots.

The constraints for the Student Grouping Problem in this study is simpler than the constraints for the Educational
Timetabling Problem. The hardest student to be scheduled is simply the student who has taken the highest number of
examinations.

The hardest first heuristics thus generates Student Grouping by greedily assigning the hardest student (student with the
most courses) to its best-fitting group (the group that results in the least penalty) with no constraint violation. When no such
group exists, a new group is created for the student. The pseudocode for the Hardest first heuristic is shown in
Algorithm \ref{algo.1}.

\begin{algorithm2e}[H]
\DontPrintSemicolon  
  \KwInput{List of students and courses offered to be grouped}
  \KwOutput{Array of student grouped into subsets}
  $students$ = list of students in decreasing order of the number of courses registered  \\
  $SG$ = Initialization of student group as an empty array \\
  \For{$i \gets 1$ \textbf{to} $length(students)$}
   {
       \If{$students[i]$ can be assigned to any group in $SG$ without any constraint violation}
           { Assign $students[i]$ to its best-fitting group in $SG$ }
       \Else
       {
         	Assign $students[i]$ to a newly created group, and add the group to $SG$
       }
   }
   \Return{$SG$}   
\caption{Pseudocode for Student Grouping by the Hardest First Ordering heuristics}\label{algo.1}
\end{algorithm2e}

\subsection{Ant Colony Optimization}
The Ant Colony Optimization (ACO) \cite{Dorigo1996} algorithm is inspired from the way ants navigate to a food source through cooperation. Ants drop pheromone trails on promising paths for other ants to follow, and if the path is still promising more pheromones are deposited on the trail, thus increasing the likelihood that other ants will follow the trail. Using this concept, an ant traverses a path to a full solution to an optimization problem. The quality of the solution is determined and then the path is updated based on the quality of the found solution. Over several iterations and ant traversals, the ACO converges to always guide ants to follow the path with the highest pheromone trail, which in turn will lead to an optimal solution.

A number of variations to the ACO exist and have been used to solve meta-heuristic  problems such as the University Course Timetabling Problem \cite{Socha2002, Kenekayoro2016}, Student Project Allocation Problem \cite{kenekayoro2020population}, Nurse Rostering Problem \cite{Ramli2020}. However, the following concepts as described in \cite{kenekayoro2020population} are the basis of all ACO algorithms. 

\textbf{Representation:} The solution search space for an optimization problem can be represented as a graph. An ant traversal from the start node to the end node forms a solution to the optimization problem. For the student grouping problem in this study, each node visited on the graph assigns a student to a group, thus the graph can be represented as an $n x m$ matrix where n rows is the number of groups and m columns is the number of students to be grouped. The number of groups n is determined by the number of groups in a solution found by the Hardest First Ordering heuristics algorithm, ensuring that final solutions cannot have more groups than the number of groups found using the Hardest First Ordering algorithm.

\textbf{Initialization}: The weight of edges on the graph representing the solution space of an optimization problem determines the probability that an ant will follow that edge. In the Max-Min ant system \cite{Socha2002} which used in this study, the maximum edge weight is $T_{max} = 10$ and the minimum edge weight is $T_{min} = 0.1$, with edges (values in the representation matrix) initialized to $T_{max}$.

\textbf{Traversal}: From an initially empty solution, as an ant traverses the solution space represented as a graph by visiting nodes, each node visited adds a student grouping to the initially empty solution to form an updated partial solution. At the end of an ant traversal, a complete solution to the student grouping problem is found. An ant decides the node that will be added to the partial solution based on a probability determined by the level of pheromone trail and if visiting that node does not result in a constraint violation. Thus, nodes whose edges have higher T values are more likely to be visited than those with lower T values. Here lies the difference between the Hardest First Ordering heuristic algorithm and the Ant Colony Optimization algorithm, the hardest first ordering heuristic simply assigns a student to its best-fitting group while the ant colony optimization assigns a student to a group based on a probability controlled by a T value that is updated as the ant colony optimization algorithm progresses over several iterations. 

\begin{algorithm2e}[H]
\DontPrintSemicolon  

  Initialize the number of ants as $numAnts$ \\
  $bestSolution$ = ant traversal of solution space to generate a solution \\
  \For{$i \gets 2$ \textbf{to} $numAnts$}
   {
       $antSolution$ = ant traversal of solution space to generate a solution \\
              \If{$quality(antSolution) < quality(bestSolution)$ }
           { $bestSolution$ = $antSolution$ }
   }
   \Return{$bestSolution$}   
\caption{Pseudocode for ants’ traversal in an iteration of the Ant Colony Optimization Algorithm}\label{algo.2}
\end{algorithm2e}

\textbf{Update}: After each iteration, the edges that form a path to the final solution are updated based on the quality of the solution. This depicts the process of ants dropping pheromone trails on promising paths, as higher T values increase the likelihood of subsequent ants following that path.  The reward determined from Equation \ref{equ.5} is added to the T values on the solution path.  In cases when the T value becomes greater than $T_{max}$ the T value is set to $T_{max}$.

\begin{equation} 
reward = \left\{\begin{matrix}
1 & if~globalBest \geq currentQuality \\ 
\frac{1}{currentQuality - globalBest} & otherwise 
\end{matrix}\right.
\label{equ.5}
\end{equation}

\textbf{Evaporation}: To avoid convergence to a local minimum the edge weights are reduced after each ant traversal. This depicts the effect of wind reducing the pheromone trails from previously promising paths not followed when a more optimal path is found. Evaporation is particularly important in the ant colony optimization algorithm in this study because there are no negative rewards as shown in Equation \ref{equ.5}, so evaporation is the only way to reduce the trail (T value) on non-optimal paths which have been initialized to $T_{max}$. T values are updated through evaporation using Equation \ref{equ.6}, where $\rho$ in the range of 0.1 to 0.9 controls the rate of evaporation. Higher evaporation rate speeds up convergence but reduces the solution space searched. In cases where the T value becomes less than $T_{min}$, the T value is set to $T_{min}$.

\begin{equation} 
T_{new} = T_{old} \times (1 - \rho)
\label{equ.6}
\end{equation}

\begin{algorithm2e}[H]
\DontPrintSemicolon  

  n = number of groups determined by the number of groups in the solution by hardest first heuristic algorithm \\
  m = number of students \\
  $T_{max}$ = 10, $T_{min}$ = 0.1, $\rho$ = 0.02, $globalBest$ = None \\ 
  SG =  $n \times m$ matrix with values initialised to $T_{max}$ \\
  \For{$i \gets 1$ \textbf{to} $numIterations$}
   {
       $cycleBest$ =  ant traversal as described in Algorithm \ref{algo.2} \\
       Reduce pheromone trail in SG by evaporation \\
       Update pheromone trail in $cycleBest's$ solution path \\
              \If{$quality(cycleBest) < quality(globalBest)$ }
           { $globalBest$ = $cycleBest$ }
   }
   \Return{$globalSolution$}   
\caption{Pseudocode of the Ant Colony Optimization for solving the Student Result Grouping Problem}\label{algo.3}
\end{algorithm2e}

Three parameters guide the Min-Max ant colony optimization algorithm to a solution, $\rho$ which controls the speed of evaporation, $\alpha$ which controls the probability that the subsequent student to be grouped is selected by hardest first heuristic and $\beta$ which controls the probability that the next student to be grouped is selected by the level pheromone trail on its path. In this study, the parameters were manually tuned to $\rho = 0.02; \alpha=0; \beta = 1$, however, as all possible values were not investigated in this study, it may be possible to achieve better solutions with different values.    

\subsection{Genetic Algorithm (GA)}
Inspired from natural evolution the Genetic Algorithm for local search \cite{Ulder1991} creates  an initial population of solutions to an optimization problem and updates the population in subsequent generations through mutation, crossover and selection operators. These genetic operators which are used in solving the student result grouping problem are described subsequently.

\textbf{Representation}: Unlike the ACO and hardest first ordering heuristic algorithms described in previous sections, final solutions are not constructed in a way such that no constraint is violated. An individual in the population as shown in Fig 1 is simply represented as a list of integers where the position i on the list represents $Student_i$ and the value in the $i^{th}$ position represents the group $Student_i$ has been assigned to.  The length of an individual is the number of students to be grouped and the number of unique values (allele) in the list of integers is the number of groups in the solution.

Heuristic information that is used in the ACO to determine the maximum number of groups is not used in the GA solution. Allele values (group a student is assigned to) is set to a random number in the range not greater than the number of students to be grouped, thus in the worst case, the final number of groups will be equal to the number of students. 

\begin{figure}[h]

{ \includegraphics[scale=0.6]{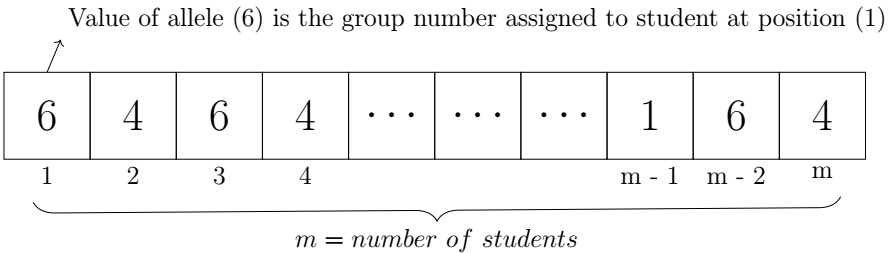} }
\caption{An individual in a GA population that represents  a solution to the Student Result Grouping Problem}\label{fig.1}
\end{figure}

As heuristic information is not used to generate the initial population of the GA, the quality of solutions is significantly worse than the first iteration of the ACO but a wider solution space is searched. 

\textbf{Evaluation}: Individuals are evaluated by the fitness function  shown in Equation \ref{equ.4}. The fitness function is especially important for the GA to guide the algorithm towards better solutions. For example, if two individuals have the same number of groups, the evaluation function ensures that the fitter individual is the individual that needs fewer modifications to result in a fewer number of total groups. 

\textbf{Crossover}: The crossover operator combines two individuals (parents) in the current population to generate two new individuals (offsprings) that will make up the population of the next generation. Single point crossover, two-point crossover, and uniform crossover are popular approaches to mate parents to produce offsprings through crossover.  Magalhães-Mendes \cite{Magalhaes-Mendes2013} compared the efficiency of different crossover approaches for a job scheduling problem, with experiments showing the single point crossover to produce the best average performance. However, in a study \cite{kenekayoro2020population} that solved the student project allocation problem, the type of crossover operator did not have any significant influence on the performance of the genetic algorithm, emphasizing that no method is guaranteed to outperform others in all problems. 

\textbf{Mutation} The mutation operator changes a single allele in an individual. In this study, the mutation operator changes the value of an allele to a random integer not greater than the total number of students to be grouped.

\textbf{Selection}: A subset of individuals from the parents (individuals in the current generation) and offsprings (determined through mutation and crossover operators) are selected to make up the population of the next generation. The python library DEAP \cite{Fortin2012} that is used in this study implements a number of selection techniques. Amongst which are tournament and roulette wheel selection methods that are widely used for single-objective genetic algorithms such as the student grouping problem in this study.

In tournament selection which is regarded as one of the most popular selection strategies for genetic algorithms \cite{Filipovic2012}
given a tournament size n, a random number of n individuals are selected from the parents and offsprings and the best individuals in each m tournaments make up the individuals in the next generation.

In roulette wheel selection, from a set of individuals (parents and offsprings), an individual is selected for the next generation by a probability proportional to its fitness. The probability of a fitter individual to be selected is higher than less fit individuals.
  
Razali and Geraghty \cite{Razali2011} has shown that tournament selection can outperform other selection techniques for some optimization problems, thus this strategy with a tournament size (n = 3), probability of mutation = 0.5 and probability of crossover = 0.5 is used in the genetic algorithm in this study.

\begin{algorithm2e}[H]
\DontPrintSemicolon  

  m = population size \\ 
  create the initial population with m number of individuals \\
  \While{stopping condition not met}
   {
       Generate offsprings by  crossover and mutation operations on parents \\
       Evaluate the fitness of offsprings \\
       Select individuals form parents and offsprings for the next generation by tournament selection whilst preserving the fittest individual \\
   }
   \Return{fittest individual in the final population}   
\caption{Pseudocode of the genetic algorithm for solving the student result grouping problem}\label{algo.4}
\end{algorithm2e}

Using the building blocks of the genetic algorithm as previously described and implemented in DEAP python library, the algorithm is run until 20 non-improving generations is reached. The pseudocode of the genetic algorithm is shown in Algorithm \ref{algo.4}. 

\section{Result and Discussion}

Table \ref{tab:5-HFO-RO-Result} and Table \ref{tab:6-ACO-GA-Result} show the result of grouping students with the Hardest First Ordering, the Genetic Algorithm, Ant Colony Optimization and the Random Ordering meta-heuristic techniques. The values in Table \ref{tab:5-HFO-RO-Result} and Table \ref{tab:6-ACO-GA-Result} indicate the quality (minimum, maximum and average) of solutions determined by Equation \ref{equ.4} after each algorithm was run 10 times.
The Hardest First Ordering heuristic is deterministic thus does not benefit from restarts as other algorithms with some degree of randomness. 
In the majority of cases, the ant colony optimization algorithm performed better than other algorithms, however,  the Hardest First Ordering heuristic can produce a good enough solution with fewer trials compared to others.
\newpage

\begin{table}[h]
\caption{Result of grouping students for the presentation of examination scores whilst satisfying the SRG problem constraints listed in Section 2 with the Hardest First Ordering (HFO) and Random Ordering (RO) meta-heuristic techniques }\label{tab:5-HFO-RO-Result}
\centering
\tiny
\begin{tabular}{lllllll} 
\hline
                  & \multicolumn{3}{c}{\textbf{HFO}}           & \multicolumn{3}{c}{\textbf{RO}}             \\
\textbf{Instance} & \textbf{Min} & \textbf{Max} & \textbf{Avg} & \textbf{Min} & \textbf{Max} & \textbf{Avg}  \\ 
\hline
RGD41107          & 95.423       & 95.423       & 95.423       & 86.44        & 171.93       & 108.66        \\
RGD4152           & 118.14       & 118.14       & 118.14       & 75.18        & 108.17       & 90.25         \\
RGD4185           & 175.34       & 175.34       & 175.34       & 128.75       & 260.89       & 183.31        \\
RGD42118          & 4.70         & 4.70         & 4.70         & 4.70         & 4.70         & 4.70          \\
RGD4263           & 191.29       & 191.29       & 191.29       & 156.53       & 249.02       & 193.43        \\
RGD4296           & 109.16       & 109.16       & 109.16       & 104.45       & 174.86       & 141.71        \\
RGD41118          & 4.25         & 4.25         & 4.25         & 4.25         & 4.25         & 4.25          \\
RGD4196           & 136.30       & 136.30       & 136.30       & 104.01       & 192.17       & 149.45        \\
RGD4241           & 147.09       & 147.09       & 147.09       & 104.50       & 194.42       & 150.24        \\
RGD4274           & 171.90       & 171.90       & 171.90       & 166.85       & 302.35       & 233.21        \\
RGD4141           & 50.19        & 50.19        & 50.19        & 33.08        & 67.81        & 49.89         \\
RGD4174           & 98.64        & 98.64        & 98.64        & 100.04       & 179.63       & 134.71        \\
RGD42107          & 65.03        & 65.03        & 65.03        & 60.88        & 115.93       & 83.16         \\
RGD4252           & 13435.17     & 13435.17     & 13435.17     & 14443.08     & 18726.77     & 16668.17      \\
RGD4285           & 176.58       & 176.58       & 176.58       & 198.65       & 287.94       & 244.73        \\
RGD4163           & 60.59        & 60.59        & 60.59        & 57.62        & 84.50        & 69.98         \\
\hline
\end{tabular}
\end{table} 

\begin{table}[h]
\caption{Result of grouping students for the presentation of examination scores whilst satisfying the SRG problem constraints listed in Section 2 with the Genetic Algorithm (GA) and the Ant Colony Optimization (ACO) meta-heuristic techniques}\label{tab:6-ACO-GA-Result}
\centering
\tiny
\begin{tabular}{lllllll} 
\hline
                  & \multicolumn{3}{c}{\textbf{GA}}            & \multicolumn{3}{c}{\textbf{ACO}}            \\
\textbf{Instance} & \textbf{Min} & \textbf{Max} & \textbf{Avg} & \textbf{Min} & \textbf{Max} & \textbf{Avg}  \\ 
\hline
RGD41107          & 89.12        & 98.69        & 92.25        & 69.76        & 91.73        & 83.61         \\
RGD4152           & 75.78        & 84.32        & 78.27        & 58.00        & 78.09        & 66.37         \\
RGD4185           & 129.15       & 164.32       & 143.53       & 127.20       & 160.73       & 134.17        \\
RGD42118          & 4.70         & 4.70         & 4.70         & 4.70         & 4.70         & 4.70          \\
RGD4263           & 125.66       & 218.49       & 175.16       & 132.36       & 161.31       & 139.11        \\
RGD4296           & 101.17       & 143.23       & 113.64       & 99.53        & 108.43       & 104.50        \\
RGD41118          & 4.25         & 4.25         & 4.25         & 4.25         & 4.25         & 4.25          \\
RGD4196           & 101.35       & 136.45       & 126.60       & 101.32       & 113.00       & 103.21        \\
RGD4241           & 123.20       & 154.45       & 134.55       & 104.26       & 115.17       & 108.99        \\
RGD4274           & 166.01       & 239.39       & 209.11       & 178.53       & 191.91       & 184.68        \\
RGD4141           & 33.08        & 51.43        & 44.58        & 33.08        & 34.10        & 33.33         \\
RGD4174           & 96.00        & 143.21       & 105.09       & 95.40        & 104.47       & 97.89         \\
RGD42107          & 60.50        & 89.86        & 72.05        & 57.97        & 59.19        & 58.40         \\
RGD4252           & 14417.77     & 14417.77     & 14417.77     & 14459.32     & 15560.84     & 14811.29      \\
RGD4285           & 158.29       & 237.22       & 196.17       & 180.56       & 240.65       & 207.61        \\
RGD4163           & 58.86        & 84.58        & 66.25        & 57.25        & 57.75        & 57.38         \\
\hline
\end{tabular}
\end{table}

The algorithms, apart from the GA are guaranteed to produce a feasible solution (all grouping conditions are met) when such a grouping exists. Partial solutions are generatively updated in such a way that a student is assigned to a group that meets all assignment conditions, and when no such group exists,  a new group is created. In the random and hardest first ordering heuristics, students are grouped greedily (assigned to the best fitting group), while in the ant colony optimization algorithm, an assignment is based on a probability controlled by the pheromone trail. Thus, the quality of these algorithms is largely dependent on the order in which students are assigned to groups. 

The GA does not update partial solutions ensuring that each update does not break feasibility but generates a complete solution randomly and then iteratively updates the solution through mutation and crossover operations until an optimal solution is found. As a result of this, the majority of individuals in the initial generations of the Genetic Algorithm are infeasible solutions.

Generating solutions for the GA is faster than other algorithms in this study as each student assigned to a group does not require an evaluation to check for feasibility, but the GA requires more iterations to converge to an optimal solution.  The fitness function is also important because even though the main aim is to have a minimal number of groups, from the fitness function, the GA should be able to identify which solution is closer to the feasibility even if they have the same number of student groups. 

Given that 26 columns (13 for new courses and 13 for old courses) are available to show student results, 13 fixed number of columns are reserved for new and old courses. In special cases such as in the RGD4252 instance, it is not possible to find a feasible solution because a student in that instance took exams for more than 13 new courses. Thus, it may be necessary to allow groups to have a flexible number of columns for new or old courses, so that in special cases columns reserved of new or old courses may be changed dynamically.

As the  ACO outperformed other algorithms as shown in Table \ref{tab:6-ACO-GA-Result}. Table \ref{tab:7-Dynamic Cols with ACO} shows the quality of solutions achieved when the number of columns for new and old courses was dynamically determined. The quality of solutions when the columns are dynamic is better than those achieved with a fixed number of columns and feasible solutions were found for all instances in the dataset used in this study.

\begin{table}[H]
\caption{Result of grouping students for presentation of results using  the Ant Colony Optimization Algorithm with dynamic number of columns }\label{tab:7-Dynamic Cols with ACO}
\tiny
\centering
\begin{tabular}{llll}
\hline
\textbf{Instance} & \textbf{Min}    &  \textbf{Max}    &  \textbf{Average}  \\
\hline
RGD41107 & 70.59  & 91.47  & 86.27    \\
RGD4152  & 59.79  & 78.89  & 71.23    \\
RGD4185  & 127.16 & 160.42 & 132.84   \\
RGD42118 & 4.70   & 4.70   & 4.70     \\
RGD4263  & 130.81 & 166.45 & 137.03   \\
RGD4296  & 98.98  & 110.66 & 103.34   \\
RGD41118 & 4.25   & 4.25   & 4.25     \\
RGD4196  & 98.98  & 110.66 & 103.34   \\
RGD4241  & 104.58 & 123.63 & 111.27   \\
RGD4274  & 183.42 & 244.98 & 198.40   \\
RGD4141  & 33.08  & 34.10  & 33.30    \\
RGD4174  & 96.11  & 104.27 & 98.39    \\
RGD42107 & 57.97  & 59.30  & 58.50    \\
RGD4252  & 453.06 & 529.19 & 495.70   \\
RGD4285  & 183.63 & 235.45 & 211.56   \\
RGD4163  & 57.25  & 58.09  & 57.33  \\
\hline
\end{tabular}
\end{table}

\section{Conclusions}
Several NP-hard problems are faced annually in Higher Education Institutions, among these problems include the timetabling and project allocation problems which have been  extensively researched. The student result grouping problem is similar to other NP-hard problems faced in HEIs as it also involves grouping students albeit with its unique constraints. Thus, techniques used in solving other NP-hard problems can be adapted to solve the student result grouping problem investigated in this study.

This study elaborately described the student result grouping problem faced in a case study Higher Education Institution and found suitable solutions to this problem by ordering heuristics (hardest first and random ordering), ants colony optimization and the genetic algorithm, demonstrating the possibility of solving this kind of problems with well-known heuristic techniques that have been used in solving other categories of NP-hard problems.

The genetic and the ant colony optimization algorithms performed better than ordering based techniques and an adaptation of the problem in the case study institution improved the quality of final solutions. Ordering based heuristics can, however, be used to find a quick good enough solution.

The parameters for the ant colony optimization and genetic algorithms were manually tuned, so better results may be achieved with more effective parameter search algorithms.

In the case study institution, the student result grouping problem was previously solved by greedily assigning or creating groups based on the order students were stored in the database. In most instances, the final solutions were worse than any of the techniques investigated in this study. Thus, research on other lesser-known combinatorial problems faced in institutions which report on adequate methods that can be used to find suitable solutions to these problems will be particularly beneficial to education software developers.\\[4mm]

\bibliographystyle{unsrt}

\end{document}